%% file: main.tex
\documentclass[english,conference, 10pt, a4paper]{IEEEtran}
\usepackage{lmodern}
\usepackage[T1]{fontenc}
\usepackage[latin9]{inputenc}
\usepackage{amstext}
\usepackage{amssymb}
\usepackage{graphicx}

\makeatletter

\providecommand{\tabularnewline}{\\}

\usepackage{hyperref}       
\usepackage{url}            
\usepackage{booktabs}       
\usepackage{amsfonts}       
\usepackage{nicefrac}       
\usepackage{microtype}      
\usepackage{times}
\usepackage{helvet}
\usepackage{courier}

\usepackage{amsbsy}
\usepackage{algorithmic}
\usepackage[english]{babel}
\usepackage{flushend}
\makeatother

\usepackage{babel}
\begin{document}

\title{Graph Memory Networks for Molecular Activity Prediction}

\author{Trang Pham, Truyen Tran, Svetha Venkatesh\\
Centre for Pattern Recognition and Data Analytics\\
Deakin University, Geelong, Australia\\
\{\textit{phtra, truyen.tran, svetha.venkatesh}\}\textit{@deakin.edu.au}}
\maketitle
\begin{abstract}
\input{abs.tex}
\end{abstract}

\global\long\def\xb{\boldsymbol{x}}
\global\long\def\yb{\boldsymbol{y}}
\global\long\def\eb{\boldsymbol{e}}
\global\long\def\zb{\boldsymbol{z}}
\global\long\def\hb{\boldsymbol{h}}
\global\long\def\ab{\boldsymbol{a}}
\global\long\def\bb{\boldsymbol{b}}
\global\long\def\cb{\boldsymbol{c}}
\global\long\def\sigmab{\boldsymbol{\sigma}}
\global\long\def\gammab{\boldsymbol{\gamma}}
\global\long\def\alphab{\boldsymbol{\alpha}}
\global\long\def\betab{\boldsymbol{\beta}}
\global\long\def\rb{\boldsymbol{r}}
\global\long\def\gb{\boldsymbol{g}}
\global\long\def\Deltab{\boldsymbol{\Delta}}
\global\long\def\wb{\boldsymbol{w}}
\global\long\def\vb{\boldsymbol{v}}
\global\long\def\eb{\boldsymbol{e}}
\global\long\def\sb{\boldsymbol{s}}
\global\long\def\ub{\boldsymbol{u}}
\global\long\def\fb{\boldsymbol{f}}
\global\long\def\mb{\boldsymbol{m}}
\global\long\def\qb{\boldsymbol{q}}

\section{Introduction}

\input{intro.tex}

\section{Related Work}

\input{related.tex}

\section{Graph Memory Networks \label{sec:Methods}}

\input{method.tex}

\section{Experiments and Results\label{sec:Results}}

\input{exp.tex}

\section{Discussion\label{sec:Discussion}}

\input{discuss.tex}

\section*{Acknowledgments}

The paper is partly supported by the Telstra-Deakin CoE in Big Data
and Machine Learning.

\bibliographystyle{IEEEtran}
\bibliography{../../bibs/ME,../../bibs/truyen,../../bibs/trang}

\end{document}

%% file: abs.tex
Molecular activity prediction is critical in drug design. Machine
learning techniques such as kernel methods and random forests have
been successful for this task. These models require fixed-size feature
vectors as input while the molecules are variable in size and structure.
As a result, fixed-size fingerprint representation is poor in handling
substructures for large molecules. Here we approach the problem through
deep neural networks as they are flexible in modeling structured data
such as grids, sequences and graphs. We train multiple BioAssays using
a multi-task learning framework, which combines information from multiple
sources to improve the performance of prediction, especially on small
datasets. We propose Graph Memory Network (GraphMem), a memory-augmented
neural network to model the graph structure in molecules. GraphMem
consists of a recurrent controller coupled with an external memory
whose cells dynamically interact and change through a multi-hop reasoning
process. Applied to the molecules, the dynamic interactions enable
an iterative refinement of the representation of molecular graphs
with multiple bond types. GraphMem is capable of jointly training
on multiple datasets by using a specific-task query fed to the controller
as an input. We demonstrate the effectiveness of the proposed model
for separately and jointly training on more than 100K measurements,
spanning across 9 BioAssay activity tests.

%% file: intro.tex
Predicting biological activities of molecules in the target environments
is a crucial step in the drug discovery pipeline. Much research has
focused on the analysis of quantitative structure-activity relationships
(QSAR), which results in a myriad of molecular descriptors \cite{cherkasov2014qsar}.
For the last 15 years, machine learning has played an important role
in the prediction pipeline, that is, mapping the molecular descriptors
into its activity classes. Successful machine learning methods are
well-established, including kernel methods \cite{burbidge2001drug,jorissen2005virtual},
random forests \cite{svetnik2003random} and gradient boosting \cite{svetnik2005boosting}.
These models take as input a fixed-size feature vector that represents
molecular properties, as known as fingerprints. The fingerprint encodes
the presence of substructures in a molecule, which are then hashed
into a fixed-size feature vector. However, the number of substructures
in large molecules might be huge, leading to many hash collisions
and information loss.

More recently, \emph{ deep learning} \cite{lecun2015deep} has started
to make impact in drug discovery \cite{baskin2016renaissance}, following
their record-breaking successes in vision and languages. The new
power comes from a mixture of better architectures (e.g., with hundreds
of layers), better training algorithms (e.g., dropout, batch normalization
and adaptive gradient descents), and faster tensor\textendash native
processors (e.g., graphic processing units). One of the initial successes
was the winning of the Merck molecular activity challenge\footnote{https://www.kaggle.com/c/MerckActivity}
in 2012 by deep neural nets \cite{dahl2014multi}. Another crucial
property of deep learning is that it is very flexible in modeling
data structures such as images, sequences and graphs. Molecular structures
can be modeled by neural networks working directly on graphs, such
as Graph Neural Network \cite{scarselli2009graph}, diffusion-CNN
\cite{atwood2016diffusion}, and Column Network \cite{pham2017column}.
Recently, there has been deep learning models successfully applied
on molecular data \cite{duvenaud2015convolutional,kearnes2016molecular,gilmer2017neural}.
Most models start with node representations by taking into account
of the neighborhood structures, typically through convolution and/or
recurrent operations. Node representations can then be aggregated
into graph representation. It is akin to representing a document\footnote{A document can be considered as a linear graph of words.}
by first embedding words into vectors (e.g., through word2vec) then
combining them (e.g., by weighted averaging using attention). We conjecture
that a better way is to learn graph representation directly and simultaneously
with node representation\footnote{This is akin to the spirit of paragraph2vec \cite{le2014distributed}.}.

We aim to efficiently \emph{learn distributed representation of graph},
that is, a map that turns variable-size graphs into fixed-size vectors
or matrices. Such a representation would benefit greatly from a powerful
pool of data manipulation tools. This rules out traditional approaches
such as graph kernels \cite{vishwanathan2010graph} and graph feature
engineering \cite{choetkiertikul2017predicting-tse}, which could
be either computation or labor intensive.

Another challenge in biological activity prediction is that the screening
process is time-consuming, limiting the number of molecules to be
tested. Traditional training of a small dataset with deep neural networks
might cause overfitting. To improve the prediction performance, work
has been proposed to use multi-task neural networks to combine data
across different biological test and learn in a multi-task scheme
\cite{dahl2014multi,ramsundar2015massively,unterthiner2015toxicity}.
These are simple feedforward neural networks that only read vector
as input features and return multiple outputs, each corresponds to
a task.

In this paper, we propose Graph Memory Network (GraphMem), a neural
architecture that generalizes a powerful recent model known as End-to-End
Memory Network \cite{sukhbaatar2015end} and apply it for modeling
 molecules. The original Memory Network consists of a controller coupled
with an unstructured and static external memory, organized as an unordered
set of cells. The controller takes a query as input and then reads
from the memory in an attentive scheme through multiple reasoning
steps before predicting an output. GraphMem, on the other hand, is
equipped with \emph{a structured dynamic memory organized as a graph
of cells}. The memory cells interact during the reasoning process,
and the memory content is refined along the way. The controller collects
information from the whole memory, so the graph representation is
embeded in the controller. The GraphMem is then applied for modeling
molecules and predicting its bioactivities as follows. First, raw
atom descriptors (or atom embedding) are loaded into memory cells,
one atom per cell, and chemical bonds dictate cell connections. A
memory cell can recurrently evolve by receiving signals from the controller
and the neighbor cells. To enable GraphMem to train in a multi-task
learning scheme, the query can be used to indicate the task index.
We validate GraphMem on more than 100K measurements, spanning across
9 BioAssay activity tests from the PubChem database\footnote{https://pubchem.ncbi.nlm.nih.gov/}.
The results demonstrate the efficacy of the proposed method against
state-of-the-arts in the field.

%% file: related.tex
\paragraph*{Memory-based neural networks}

End to End Memory Network (E2E MemNet) \cite{sukhbaatar2015end} is
a Recurrent Neural Network that has an external memory. The memory
contains multiple cells, each cell corresponds to an input vector.
The controller reads a query and repeatedly reads from the memory
before predicting an output. For the question answering task, the
query is a question and each memory cell is an input sentence or a
fact. With signals received from the query, the controller attentively
chooses appropriate information from the memory to produce the output.
A similarity architecture is Neural Turing Machine (NTM)  \cite{graves2014neural},
which also uses a controller to attentively read from a continuous
memory. Besides the read head like in an E2N MemNet, NTM has a write
head so that the memory of NTM can be erased overtime and updated
with new input. Little work has been done using the structured memory
\cite{parisotto2017neural,bansal2017relnet}. Our GraphMem differs
from these models by a dynamic memory organized as a graph of cells.
The cells interact not only with the controller but also with other
cells to embed the substructure in their states.

\paragraph*{Graph representation}

There has been a sizable rise of learning graph representation in
the past few years \cite{bronstein2016geometric,bruna2014spectral,henaff2015deep,johnson2017learning,li2016gated,niepert2016learning,schlichtkrull2017modeling}.
A number of works derive shallow embedding methods such as node2vec
and subgraph2vec, possibly inspired by the success of embedding in
linear-chain text (word2vec and paragraph2vec). Deep spectral methods
have been introduced for graphs of a given adjacency matrix \cite{bruna2014spectral},
whereas we allow arbitrary graph structures, one per graph. Several
other methods extend convolutional operations to irregular local neighborhoods
\cite{atwood2016diffusion,niepert2016learning,pham2017column}. Yet
recurrent nets are also employed along the random walk from a node
\cite{scarselli2009graph}. Our application to chemical compound classification
bears some similarity to the work of \cite{duvenaud2015convolutional},
where graph embedding is also collected from node embedding at each
layer and refined iteratively from the bottom to the top layers. However,
our treatment is more principled and more widely applicable to multi-typed
edges.

\paragraph*{Multi task neural network for molecular activity prediction}

An application of neural networks for molecular activity prediction
is to train a neural network on an assay toward a specific test. This
method can fit and predict the data well when the training data is
sufficient. However, molecular activity tests are costly, hence, the
datasets are normally small, causing overfitting on neural network
models. To improve the prediction performance, multiple tests can
be jointly trained by a single neural network. The model for multi-task
learning is still a neural network that reads the input feature vector,
but there is a separated output for each task. This model has been
applied successful in work for molecular activity prediction such
as QSAR prediction \cite{dahl2014multi}, massive drug discovery \cite{ramsundar2015massively}
with a large dataset of 40M measurements over more than 200 tests
and toxicity prediction challenge \cite{unterthiner2015toxicity}.

%% file: method.tex
In this section, we present Graph Memory Networks (GraphMem) for general
graphs. An illustration is given in Fig.~\ref{fig:GraphMem}.

\subsection*{Definition and notation: Multi-relational graphs}

A graph is a tuple $\text{\textbf{G}=\{\textbf{A}, \textbf{R}, \textbf{X}\}}$,
where $\text{\textbf{A}}=\left\{ a^{1},...,a^{N}\right\} $ are $M$
nodes. $\text{\textbf{X}}=\left\{ \xb^{1},...,\xb^{M}\right\} $ is
the set of node features, where $\xb^{i}$ is the feature vector of
node $a^{i}$. $\text{\textbf{R}}$ is the set of relations in the
graph. Each tuple $\left\{ a^{i},a^{j},r,\bb^{ij}\right\} \in\text{\textbf{R}}$
describes a relation of type $r$ ($r=1...R$) between two nodes $a^{i}$
and $a^{j}$. The relations can be one-directional or bi-directional.
The vector $\bb^{ij}$ represents the link features. Node $a^{j}$
is a neighbor of $a^{i}$ if there is a connection between the two
nodes. Let $\mathcal{N}(i)$ be the set of all neighbors of $a^{i}$
and $\mathcal{N}_{r}(i)$ be the set of neighbors connected to $a^{i}$
through type $r$. This implies $\mathcal{N}(i)=\cup_{r}\mathcal{N}_{r}(i)$.

\subsection*{Model overview}

\begin{figure}[h]
\centering{}\includegraphics[bb=300bp 180bp 680bp 410bp,clip,width=0.5\textwidth]{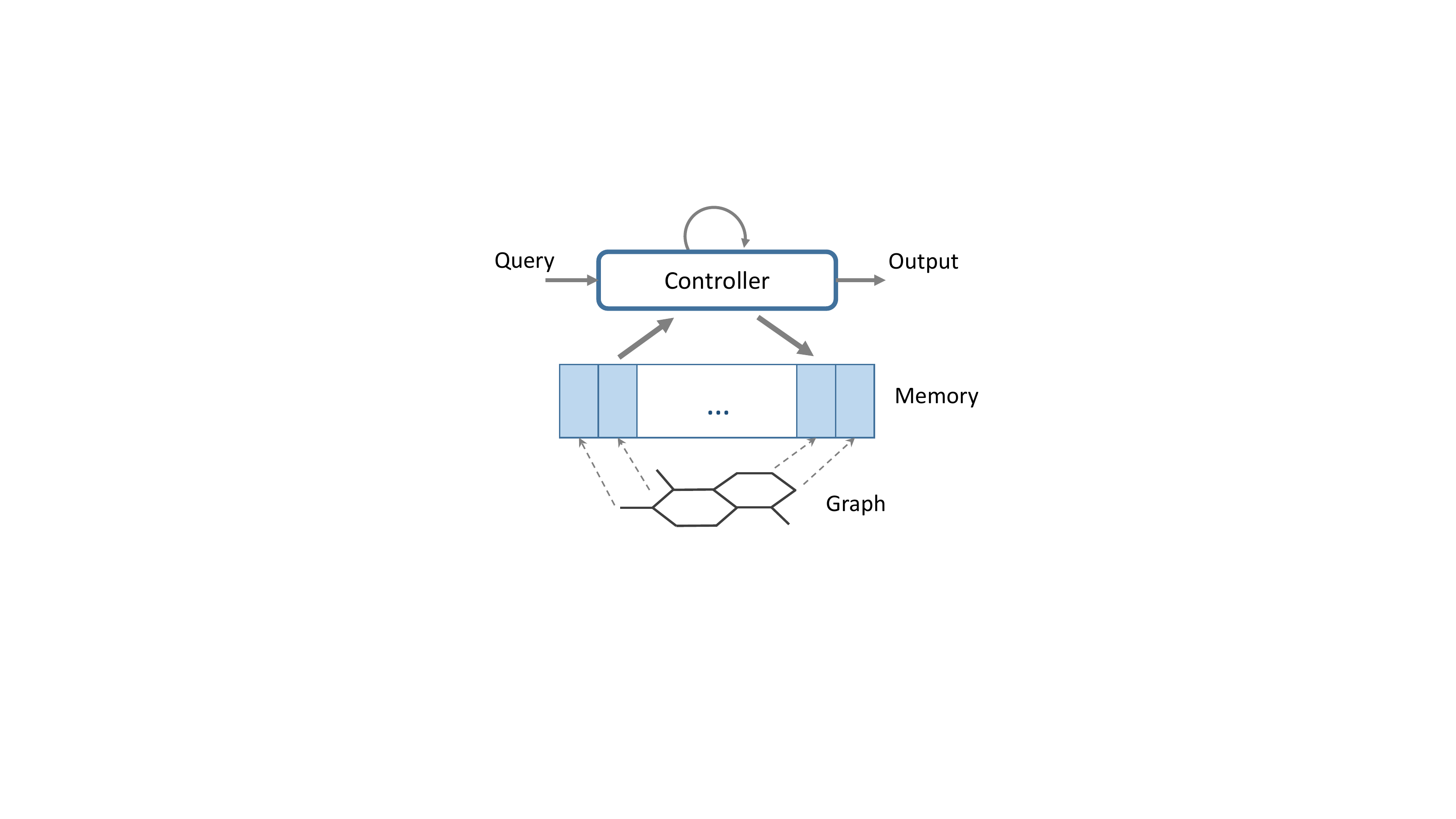}\caption{Graph Memory Network. At the first step, the controller reads the
query; the memory is initialized by the input graph, one node embedding
per memory cell. Then during the reasoning process, the controller
iteratively reads from and writes to the memory. Finally, the controller
computes the output. \label{fig:GraphMem}}
\end{figure}

GraphMem (Fig.~\ref{fig:GraphMem}) consists of a controller and
an external memory, both of which, when rolled out, are recurrent
neural networks (RNNs) interacting with each other. Different from
the standard RNNs, the memory is a matrix RNN \cite{do2017learn},
where the hidden states are matrices with a graph imposed on columns.
The controller first takes the query as the input and repeatedly reads
from the memory using an attention mechanism, processes and sends
the signals back to the memory cells. Each memory cell is first initialized
by the input graph embedding, one node per cell. Then at each reasoning
step, the cell content is updated by the signals from the controller
and its neighbor memory cells in the previous step. Through multiple
steps of reasoning, the memory cells are evolved from the original
input to a refined stage, preparing the controller for generating
the output. The query setting is flexible as it has been demonstrated
in question-answering tasks \cite{sukhbaatar2015end}.

\subsection*{The controller and attentive reading}

Let $\qb$ be the query vector and $\hb_{t}$ be the state of the
controller at time $t$ $\left(t=0,...,T\right)$. First, the controller
reads the query: $\hb_{0}=g\left(W_{q}\qb\right)$. All biases are
omitted for clarity. 

During the multi-hop reasoning process to answer the query, the controller
reads the summation vector $\mb_{t}$ from the memory and updates
its state as follows:

\begin{equation}
\hb_{t}=g\left(W_{h}\hb_{t-1}+U_{h}\mb_{t}\right)\label{eq:controller_hidden}
\end{equation}

As $\mb_{t}$ is the summation of the memory, the representation of
the whole graph is embedded in $\mb_{t}$, thus embedded in the controller
$\hb_{t}$. The hidden state of the controller contains both the graph
and the query representations, which are neccessary to produce an
output. The controller predicts an output after $T$ steps of the
process of reasoning and updating. The output can be of any type corresponding
to the query. For example, the query-output pairs can be (if a graph
has a specific property - binary output) and (what is the solubility
of a molecule compound - continuous output).

To read the vector $\mb_{t}$ from the memory, a content-based addressing
scheme, also known as soft attention, is employed. At each time step
$t$ $\left(t=1,...,T\right)$, $\mb_{t}$ is a sum of all memory
cells, weighted by the probability $p_{t}^{i}$, for each memory cell
$i=1,...M$:

\begin{eqnarray*}
\ab_{t}^{i} & = & \text{tanh}\left(W_{a}\mb_{t-1}^{i}+U_{a}\hb_{t-1}\right)\\
p_{t}^{i} & = & \text{softmax}\left(\vb^{\top}\ab_{t}^{i}\right)\\
\mb_{t} & = & \sum_{i}p_{t}^{i}\mb_{t-1}^{i}
\end{eqnarray*}
where $\ab_{t}^{i}$ integrates information stored in the memory cell
$\mb_{t-1}^{i}$ and the controller state $\hb_{t-1}$, and $\vb$
is a parameter vector used to measure the contribution of memory cells
to the summation vector. With this attention mechanism, the controller
can selectively choose important nodes toward the query and the predictive
output, rather than considering them equally. The query acts like
an attention signal that guide the controller where to put more weight
on.

\subsection*{Graph-structured Multi-relational Memory}

The memory is extended from the unstructured memory in the E2E MemNet
to a graph-structured multi-relational one. Each node in the graph
$a^{i}$ has a feature vector (either extracted from the data, or
through embedding) $\xb^{i}\in\mathbb{R}^{K_{x}}$ ($i=1...M)$. The
memory consists of $M$ memory cells, each cell $\mb_{t}^{i}\in\mathbb{R}^{K_{m}}$
stores the information of the node $a^{i}$. The memory cells are
initialized by a transformation of the feature vectors: $\mb_{1}^{i}=g\left(\xb^{i}\right)$.
The memory cells connect to each other based on the node connections
in the graph. If two nodes are connected through a relation, their
corresponding memory cells have a connection. This enables the memory
cells to embed the substructures of the graph by updating their content
by the information from their neighbors.

At step $t$, each memory cell is updated by a function of the previous
memory, a write content from the controller and the memories from
the neighboring cells: $\mb_{t}^{i}=f\left(\mb_{t-1}^{i},\hb_{t},\left(\mb_{t-1}^{j}\right)_{j\in\mathcal{N}(i)}\right)$.
In our experiments, the implementation of the memory update is as
follow:

\begin{eqnarray}
\mb_{t}^{i} & = & g\left(W_{m}\mb_{t-1}^{i}+U_{m}\hb_{t}+\sum_{r}V_{r}\cb_{tr}^{i}\right)\label{eq:memory_cell_hidden}\\
\cb_{tr}^{i} & = & \sum_{j\in\mathcal{N}_{r}(i)}p^{j}\left[\mb_{t-1}^{j},\bb^{ij}\right]\label{eq:get_neighbors}
\end{eqnarray}
where $\mathcal{N}_{r}(i)$ is the neighbor of the node $a^{i}$ with
the relation type $r$ and $\cb_{tr}^{i}$ denotes the neighboring
context of relation $r$. $\left[\mb_{t-1}^{j},\bb^{ij}\right]$ is
the concatenation of the memory cell $\mb_{t-1}^{j}$ and the link
feature vector $\bb^{ij}$. $p^{j}$ is the weight of the node $a^{j}$,
which indicates how important the node $a^{j}$ toward $a^{i}$ and
$\sum_{j\in\mathcal{N}_{r}(i)}p^{j}=1$. $p^{j}$ can be learned similar
to the memory cell probabilities in the attentive reading or can be
pre-computed.

This update allows each memory cell to embed the neighbor information
in its representation, thus, capturing the graph structure information.
The neighboring update can be found in different graph-based neural
networks \cite{scarselli2009graph,li2015gated,pham2017column}. The
common idea is that each node can embed the graph substructure information
around it by iteratively updating signals from its neighbors through
multiple steps. For example, we have ($a^{1}$, $a^{2}$) and ($a^{2}$,
$a^{3}$) are two connections. At first, the memory cell $\mb_{1}^{1}$
contains the signals from $a^{2}$ and $\mb_{1}^{2}$ contains the
signals from $a^{3}$. At the second step, $\mb_{2}^{1}$ updates
signals from $\mb_{1}^{2}$, which already contains information of
$a^{3}$. If the number of steps is large enough, a node can contain
information of the whole graph.

\subsection*{Recurrent skip-connections}

The controller can be implemented in several ways. It could be a feedforward
network or a recurrent network such as LSTM. In case of feedforward
net, the query information is propagated through the memory via memory
update. In case of recurrent nets, the query information is also propagated
through the internal state of the controller. For simplicity, in this
paper, we implement the controller and the memory updates using skip-connections
\cite{pham2016faster,srivastava2015training}

\[
\zb_{t}=\alphab*\tilde{\zb}_{t}+\left(1-\alphab\right)*\zb_{t-1}
\]
where $\alphab$ is a sigmoid gate moderating the amount of information
flowing from the previous step, $\zb_{t-1}$ is the state from the
previous step and $\tilde{\zb}_{t}$ is a proposal of the new state
which is typically implemented as a nonlinear function of $\zb_{t-1}$. 

The controller $\hb_{t}$ and the memory cell $\mb_{t}^{i}$ are updated
in a fashion similar to that of $\zb_{t}$ while $\tilde{\hb}_{t}$
and $\tilde{\mb}_{t}^{i}$ are computed as in Eq.~(\ref{eq:controller_hidden}~and~\ref{eq:memory_cell_hidden}).
This makes the memory cell update similar to the one in Differential
Neural Computer \cite{graves2016hybrid}, where the memory cells are
partially erased and updated with new information. 

\textbf{Remark:} With this choice of recurrence, the entire network
can be considered as $M+1$ RNNs interacting following the structure
defined by the multi-relational graph.

\subsection*{GraphMem for multi-task learning}

GraphMem can be easily applied for multi-task learning. Suppose that
the dataset contains $n$ tasks, each task is a set of graphs toward
a specific type of output. We can use the query to indicate the task.
If a graph is from task $k$, the query for the graph is a one-hot
vector of size $n$: $\qb=[0,0,...,1,0,...]$, where $\qb^{k}=1$
and $\qb^{j}=0$ for $j=1,...,n,\text{ }j\neq k$. The task index
now becomes the input signal for GraphMem. With the signal from the
task-specific query, the attention can identify which substructure
is important for a specific task to attend on.

%% file: exp.tex
\subsection{Datasets \label{subsec:Feature-NCI}}

We conducted experiments on 9 NCI \textbf{BioAssay} activity tests
collected from the PubChem website \footnote{https://pubchem.ncbi.nlm.nih.gov/}.
7 of them are activity tests of chemical compounds against different
types of cancer: breast, colon, leukemia, lung, melanoma, central
nerve system and renal. The others are AIDS antiviral assay and Yeast
anticancer drug screen. Each BioAssay test contains records of activities
for chemical compounds. We chose the 2 most common activities for
classification: ``active'' and ``inactive''. Each compound molecule
is represented as a graph, where nodes are atoms and edges are bonds
between them. The statistics of data is reported in Table~\ref{tab:data-statistics}.
The datasets are listed by the ascending order of number of active
compounds. ``\# Graph'' is the number of graphs and ``\# Active''
is the number of active graph against a BioAssay test These datasets
are unbalanced, therefore ``inactive'' compounds are randomly removed
so that the Yeast Anticancer dataset has 25,000 graphs and each of
the other datasets has 10,000 graphs.

\begin{table}[h]
\centering{}\caption{Summary of 9 NCI BioAssay datasets.\label{tab:data-statistics}}
\begin{tabular}{cccc}
\hline 
No. & Dataset & \# Active & \# Graph\tabularnewline
\hline 
1 & AIDS Antiviral & 1513 & 41,595\tabularnewline
2 & Renal Cancer & 2,325 & 41,560\tabularnewline
3 & Central Nervous System  & 2,430 & 42,473\tabularnewline
4 & Breast Cancer & 2,490 & 29,117\tabularnewline
5 & Melanoma & 2,767 & 39,737\tabularnewline
6 & Colon Cancer & 2,766 & 42,130\tabularnewline
7 & Lung Cancer & 3,026 & 38,588\tabularnewline
8 & Leukemia & 3,681 & 38,933\tabularnewline
9 & Yeast Anticancer & 10,090 & 86,130\tabularnewline
\hline 
\end{tabular}
\end{table}

\paragraph*{Fingerprint feature extraction }

Fingerprints are the encoding of the graph structure of the molecules
by a vector of binary digits, each presents the presence or absence
of particular substructures in the molecules. There are different
algorithms to achieve molecular fingerprints and the state of the
art is the extended-connectivity circular fingerprint (ECFP) \cite{rogers2010extended}.
We use the RDKit toolkit to extraction circular fingerprints\footnote{http://www.rdkit.org/}.
The dimension of the fingerprint features is set by 1024.

\paragraph*{Graph extraction}

We also use RDKit to extract the structure of molecules, the atom
and the bond features. An atom feature vector is the concatenation
of the one-hot vector of the atom and other features such as atom
degree and number of H atoms attached. We also make use of bond features
such as bond type and a binary value indicating if a bond in a ring.

\subsection{Experiment settings}

To evaluate the benefit of multi-task learning, we trained the 9 datatasets
separately and jointly on both fingerprint features and graph structure.
In the multi-task setting, each dataset is a single task.

In the separated training setting, three common classifiers: Support
Vector Machine (SVM), Random Forest (RF) and Gradient Boosting Machine
(GMB) are trained on fingerprint features, and Neural Fingerprint
(NeuralFP) \cite{duvenaud2015convolutional} and our GraphMem are
trained on graph structure. The query of GraphMem for each dataset
is set by a constant vector.

In the joint training setting, we trained Multitask Neural Network
(MT-NN) \cite{ramsundar2015massively} on fingerprint features and
our model (MT-GraphMem) on graph structure. The query for a molecule
of task $i$ is a one-hot vector $\qb\in\mathbb{R}^{K}$, where $K$
is the number of tasks, $\qb^{i}=1$ and $\qb^{j}=0$ for all $j\neq i$.
For simplicity, we set the weights in Eq.~\ref{eq:get_neighbors}
uniformly.

For training neural networks, the training minimizes the cross-entropy
loss in an end-to-end fashion. We use ReLU units for all steps and
Dropout \cite{srivastava2014dropout} is applied at the first and
the last steps of the controller and the memory cells. We set the
number of hops by $T=10$ and other hyper-parameters are tuned on
the validation dataset.

\subsection{The impact of more datasets}

We evaluate how performance of GraphMem on a particular dataset is
affected by the increasing number of tasks. We chose AIDS antiviral,
Breast Cancer and Colon Cancer as the experimental datasets. For each
experimental dataset, we start to train it and then repeatedly add
a new task and retrain the model. The orders of the first three new
tasks are: (AIDS, Breast, Colon) for AIDS antiviral dataset, (Breast,
AIDS, Colon) for Breast Cancer dataset and (Colon, AIDS, Breast) for
Colon Cancer dataset. The orders of the remaining tasks are the same
for 3 datasets: (Leukemia, Lung, Melanoma, Nerve, Renal and Yeast).

\begin{figure*}
\centering{}%
\begin{tabular}{cc}
\includegraphics[bb=10bp 0bp 540bp 420bp,clip,width=0.45\textwidth]{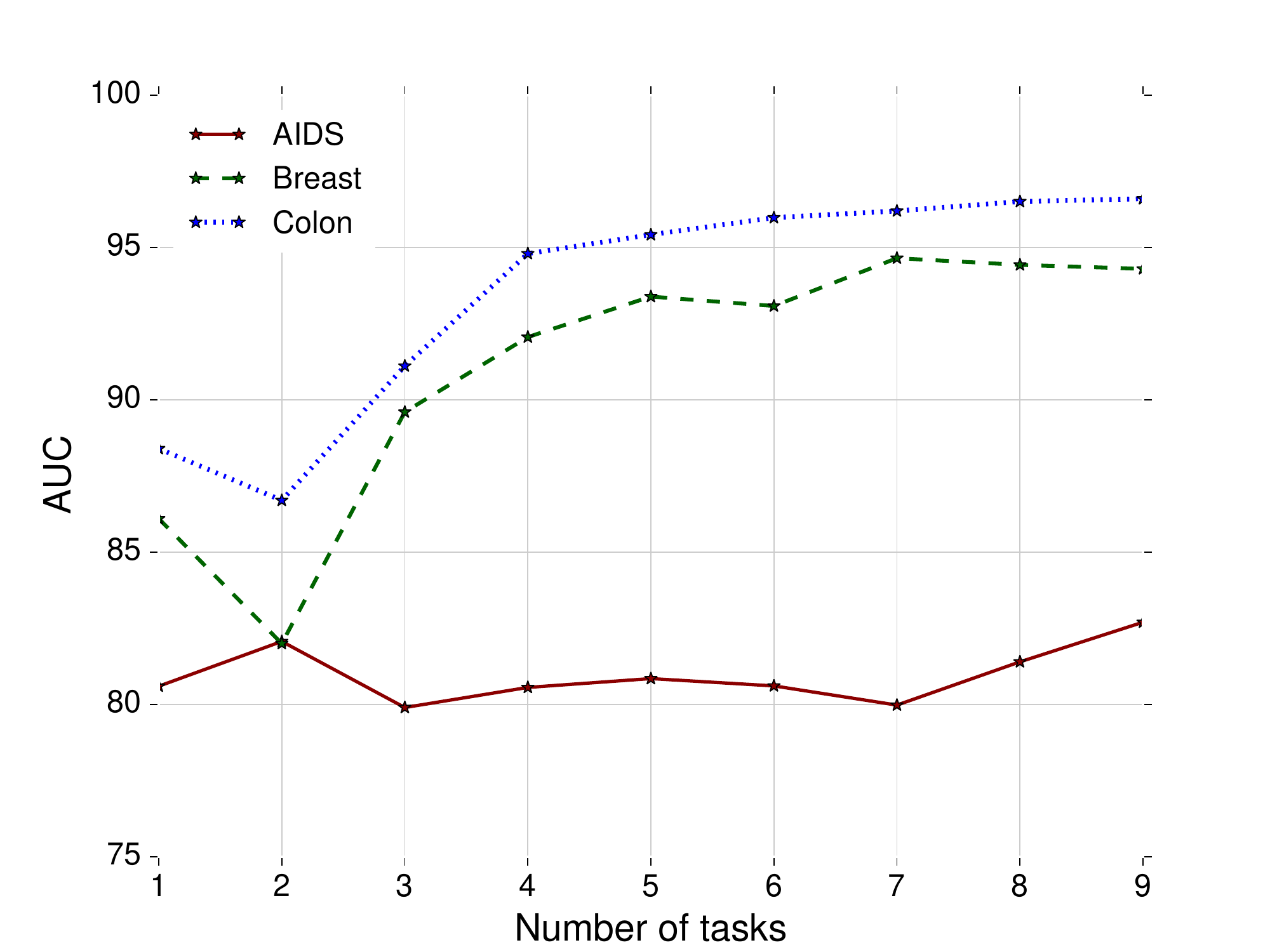} & \includegraphics[bb=10bp 0bp 540bp 420bp,clip,width=0.45\textwidth]{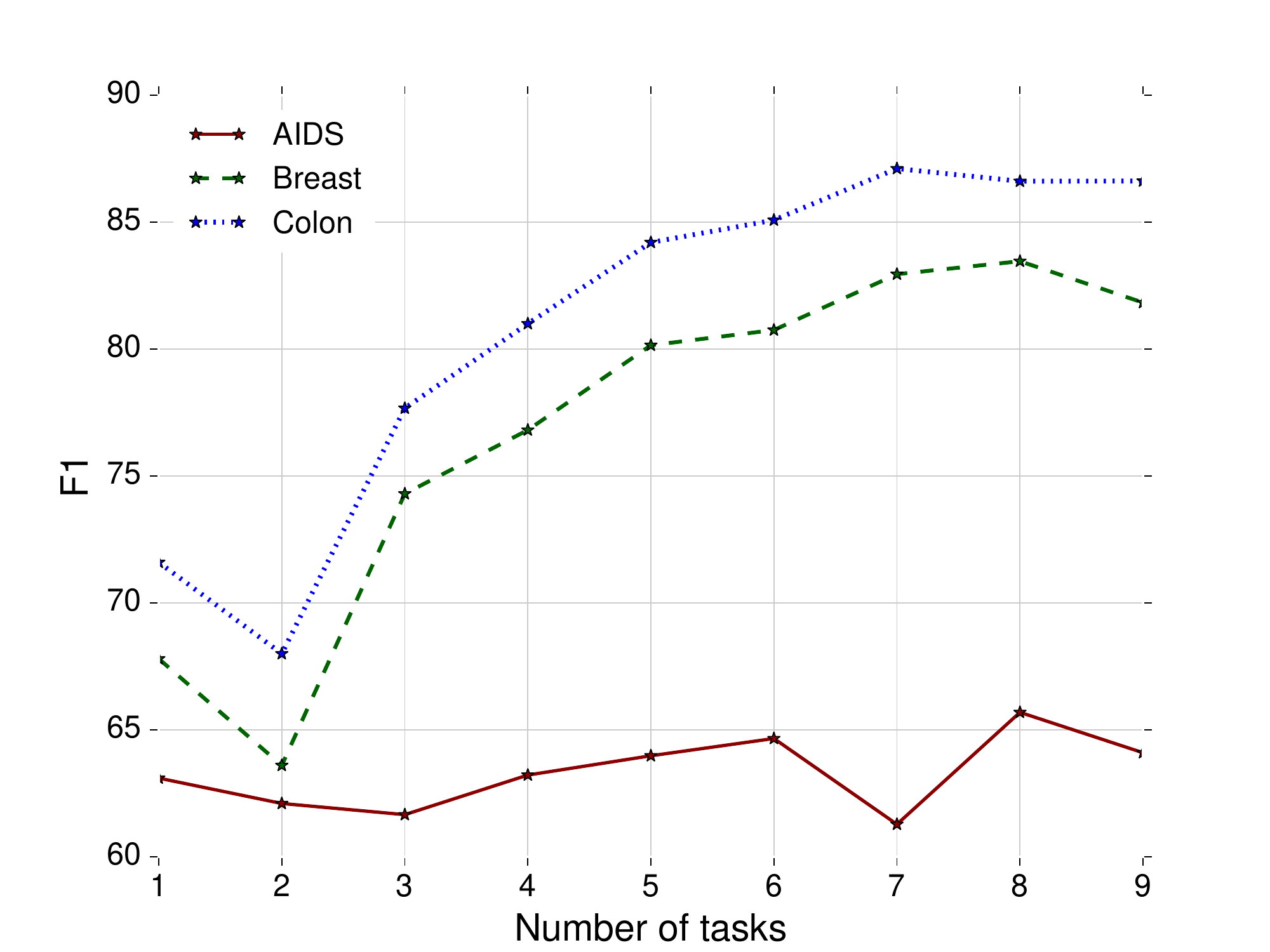}\tabularnewline
(a) & (b)\tabularnewline
\end{tabular}\caption{The performance of three datasets when increasing the number of jointly
training tasks, reported in (a) AUC and (b) F1-score.\label{fig:increasing_n_tasks}}
\end{figure*}

Fig.~\ref{fig:increasing_n_tasks} illustrates the performance of
the three chosen datasets with different number of jointly training
tasks. The performance of Breast and Colon Cancer datasets decreases
when jointly trained with AIDS antiviral task and then increases after
adding more tasks and remain steady or slightly reduce after 7 tasks.
Jointly training does not really improve the performance on the AIDS
antiviral dataset.

\subsection{Results}

\begin{table}
\centering{}\caption{Performance over all datasets, measured in Micro F1, Macro F1 and
the average AUC. \label{tab:Results-all}}
\begin{tabular}{cccc}
\hline 
Model & MicroF1 & MacroF1 & Average AUC\tabularnewline
\hline 
SVM & 66.4 & 67.9 & 85.1\tabularnewline
RF & 65.6 & 66.4 & 84.7\tabularnewline
GB & 65.8 & 66.9 & 83.7\tabularnewline
\hline 
NeuralFP \cite{duvenaud2015convolutional} & 68.2 & 67.6 & 85.9\tabularnewline
GraphMem & 69.1 & 68.7 & 85.9\tabularnewline
\hline 
MT-NN \cite{ramsundar2015massively} & \emph{75.5} & \emph{78.6} & \emph{90.4}\tabularnewline
MT-GraphMem & \textbf{77.8} & \textbf{80.3} & \textbf{92.1}\tabularnewline
\hline 
\end{tabular}
\end{table}

Table~\ref{tab:Results-all} reports results, measured in Micro F1-score,
Macro F1-score and the average AUC over all datasets. The best method
for separated training on fingerprint features is SVM with 66.4\%
of Micro F1-score and on graph structure is GraphMem with the improvement
of 2.7\% over the non-structured classifiers. The joint learning settings
improve a huge gap with 9.1\% of Micro F1-score and 10.7 \% of Macro
F1-score gain on fingerprint features and 8.7\% of Micro F1-score
and 11.6\% of Macro F1-score gain on graph structure.

\begin{figure}[t]
\centering{}\includegraphics[bb=40bp 0bp 655bp 470bp,clip,width=0.5\textwidth]{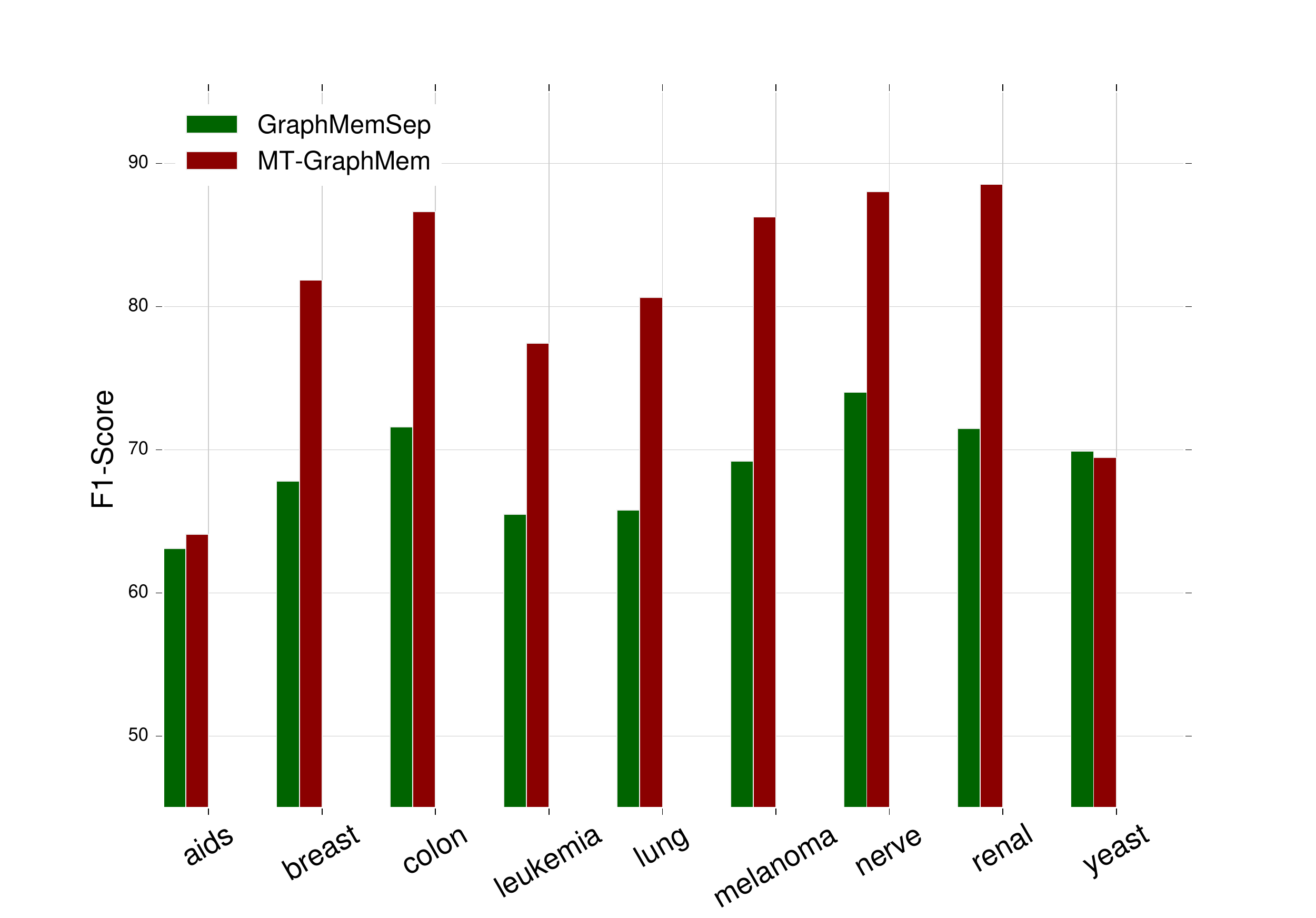}\caption{The comparison in performance of GraphMem when training separately
(GraphMemSep) and jointly (MT-GraphMem) for all datasets. Best view
in color.\label{fig:F1-score-each-dataset}}
\end{figure}

To investigate more on how multi-task learning impacts the performance
of each task, we reports the F1-score of GraphMem in both separately
and jointly training settings on each of the 9 datasets (Fig.~\ref{fig:F1-score-each-dataset}).
Joint training with GraphMem model does not improve the performance
on AIDS antiviral and Yeast anticancer datasets while for 7 datasets
on different types of cancers, joint training improves the performance
from 10\%-20\% on each task.

%% file: discuss.tex
We have proposed Graph Memory Network (GraphMem), a neural network
augmented with a dynamic and graph-structured memory and applied it
for modeling molecules. Experiments on 9 BioAssay activity tests demonstrated
that GraphMem is effective in answering queries about bioactivities
of large molecules given only the molecular graphs. We applied the
model for multi-task learning with the query indicating the task number.
However, the query is very flexible, it can be any question about
the property of a molecule.

There is room for further investigations. First, we wish to emphasize
that GraphMem is a general models for answering any query about graph
data. This opens up new applied opportunities in other domains, for
example, textual and visual question answering about interacting actors
and objects. Second, BioAssay activity ground truths used in training
for each target (e.g., a disease) are expensive to establish. We can
leverage the strength of statistics from the existing large datasets
to improve over the smaller datasets. For example, each BioAssay test
can be considered as a task and the model can jointly learn all tasks.
The task ID and other information of the molecule can be embedded
in the query. Furthermore, the memory structure in GraphMem, once
constructed from data graphs, is then fixed even though the content
of the memory changes during the reasoning process. A future work
would be deriving dynamic memory graphs that evolve with time.